\documentclass[runningheads]{llncs}

\usepackage{eccv}

\usepackage{eccvabbrv}

\usepackage{graphicx}
\usepackage{booktabs}
\usepackage{multirow}
\usepackage{todonotes}
\usepackage{pifont}

\usepackage[accsupp]{axessibility}  %

\usepackage{hyperref}

\usepackage{orcidlink}
\usepackage[T1]{fontenc}

\begin{document}
\newcommand{\xmark}{\ding{55}}
\newcommand{\tick}{\ding{51}}
\title{Depth Any Canopy: Leveraging Depth Foundation Models for Canopy Height Estimation}

\titlerunning{Depth Any Canopy}

\author{
Daniele Rege Cambrin\inst{1}
\orcidID{0000-0002-5067-2118} \and \\
Isaac Corley\inst{2}\orcidID{0000-0002-9273-7303} \and \\
Paolo Garza\inst{1}\orcidID{0000-0002-1263-7522}
}

\authorrunning{D.~Cambrin et al.}

\institute{Politecnico di Torino, Torino Italy \\
\email{\{daniele.regecambrin,paolo.garza\}@polito.it}\\
\and
University of Texas at San Antonio, San Antonio, TX USA\\
\email{isaac.corley@utsa.edu}}

\maketitle

\begin{abstract}
    Estimating global tree canopy height is crucial for forest conservation and climate change applications. However, capturing high-resolution ground truth canopy height using LiDAR is expensive and not available globally. An efficient alternative is to train a canopy height estimator to operate on single-view remotely sensed imagery. The primary obstacle to this approach is that these methods require significant training data to generalize well globally and across uncommon edge cases. Recent monocular depth estimation foundation models have show strong zero-shot performance even for complex scenes. In this paper we leverage the representations learned by these models to transfer to the remote sensing domain for measuring canopy height. Our findings suggest that our proposed Depth Any Canopy, the result of fine-tuning the Depth Anything v2 model for canopy height estimation, provides a performant and efficient solution, surpassing the current state-of-the-art with superior or comparable performance using only a fraction of the computational resources and parameters. Furthermore, our approach requires less than \$1.30 in compute and results in an estimated carbon footprint of 0.14 kgCO$_2$. Code, experimental results, and model checkpoints are openly available at %
    \href{https://github.com/DarthReca/depth-any-canopy}{github.com/DarthReca/depth-any-canopy}.
\end{abstract}

\section{Introduction}
\label{sec:intro}

\begin{figure}[t!]
    \centering
    \includegraphics[width=0.8\linewidth]{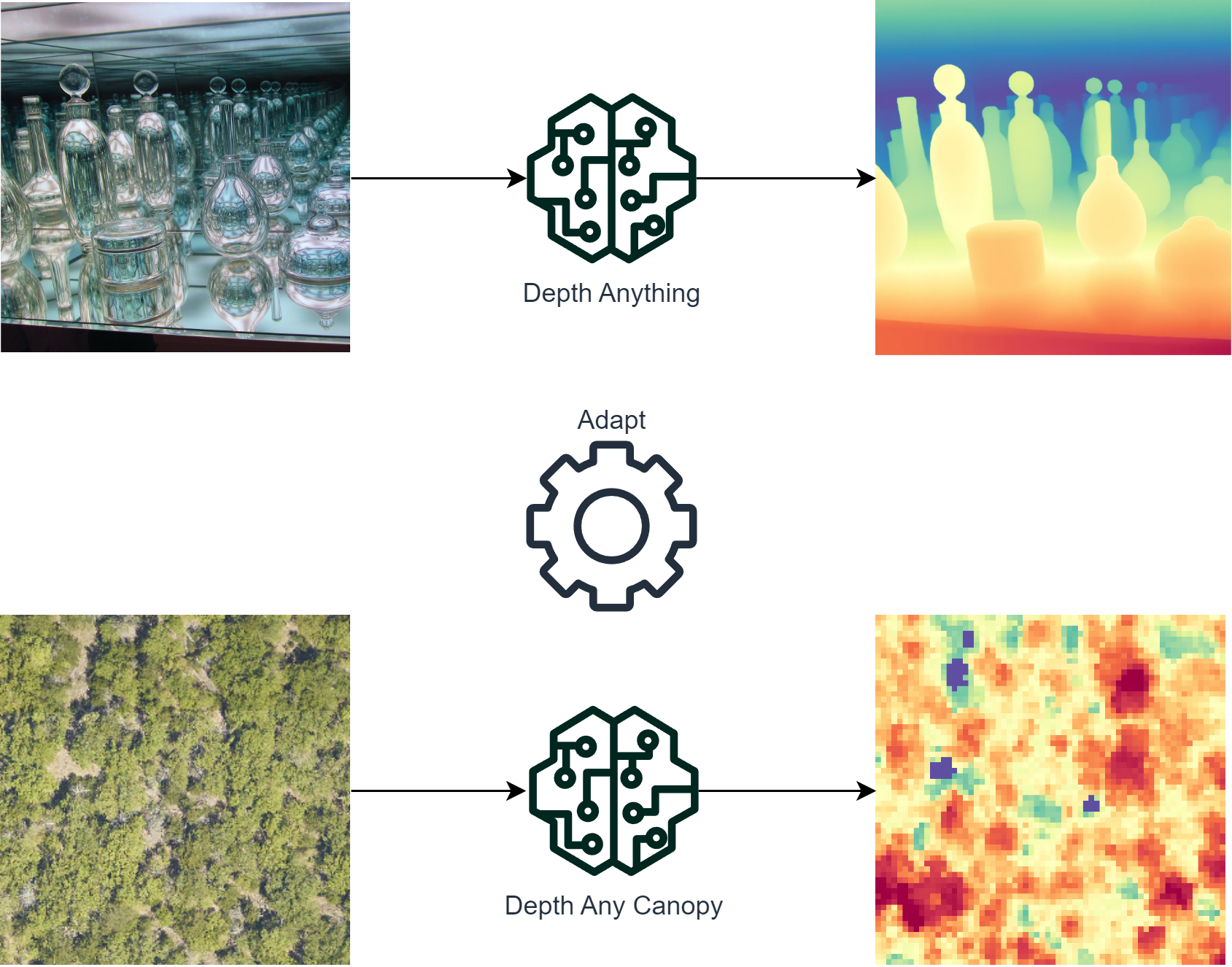}
    \caption{\textbf{From Depth Anything~\cite{depth_anything_v2} to Depth Any Canopy}. Depth Anything is a monocular depth estimation foundation model trained on natural imagery. We fine-tune and adapt Depth Anything v2 for the task of estimating tree canopy height in remote sensing imagery, resulting in Depth Any Canopy (DAC).}
    \label{fig:fig-architecture}
\end{figure}

Measuring tree canopy extent and height accurately is crucial to tracking the health of our world's ecosystem~\cite{watch2002global}. However, manual in-situ measurements or airborne laser scans (ALS) using LiDAR~\cite{andersen2001automated, gaveau2003quantifying} are expensive and slow, making them unscalable for acquiring global measurements~\cite{tomppo2010national}. Furthermore, each acquisition method comes with their own sensitivities and errors~\cite{sexton2009comparison}. 
Consequently, there is a growing need for more accessible and cost-effective approaches to canopy height estimation. 

Remotely sensed satellite data is currently in abundance and an efficient alternative; however, satellite sensors built for measuring canopy height, such as GEDI~\cite{dubayah2022gedi}, provide measurements which lack high spatial resolution. With this said, the most reasonable automated solution is to utilize machine learning to extract canopy height learned from satellite imagery geospatially aligned with ALS-derived canopy height maps~\cite{fogel2024opencanopycountryscalebenchmarkcanopy}.

One promising alternative is to develop a canopy height estimator that uses single-view remotely sensed image. However, this approach faces the substantial challenge of requiring extensive training data to achieve generalization across diverse global environments.  
Foundation models have displayed the importance of large-scale pretraining, which allows for the high-quality pretrained weights to be fine-tuned in a low-cost manner for various downstream tasks~\cite{foundation_remote_sensing,foundation_history}. 

The task of canopy height estimation from remote sensing imagery can be posed similarly to the depth estimation task, where the camera view is always fixed overhead and we seek to measure distance above ground, with ground represented as zero.
Monocular depth estimation for natural imagery has recently experienced significant zero-shot performance improvements by pretraining on complex synthetic data and then pseudo-labeling a large-scale dataset for further semi-supervised training~\cite{depth_anything_v1}. These models have shown the capability of being adapted for various applications, suggesting the opportunity for cross-domain transfer learning.

In this work, we explore the adaption of depth foundation models to remote sensing domain, specifically for tree canopy height estimation, without requiring significant amount of pretraining on remote sensing imagery. We propose leveraging Depth Anything v2~\cite{depth_anything_v2}, a state-of-the-art monocular depth estimation model, to enhance the accuracy and efficiency of canopy height estimation from aerial imagery. 

Our findings indicate that a proper finetuning of the model not only surpasses the current state-of-the-art performance but does so with fewer computational resources. Using this new model, Depth Any Canopy (DAC), we aim to provide a scalable and efficient solution for global canopy height estimation, making a step forward in providing high-quality canopy height data more accessible and contributing to better forest management and climate change mitigation efforts.

Our contributions can be summarized as follows:
\begin{enumerate}
    \item We adapt monocular depth estimation foundation models, Depth Anything v2, fine-tuning them for canopy height estimation derived from remote sensing imagery, resulting in Depth Any Canopy (DAC).
    \item We achieve superior or comparable results to a larger baseline with significant advantages of being pretrained on millions of in-domain satellite imagery.
    \item Our resulting fine-tuning process and models are efficient and low-cost. They can be reproducible using minimal consumer GPU hardware.
\end{enumerate}

The rest of the article is structured as follows. \Cref{sec:rl} discusses the related works in canopy height estimation and foundational models in computer vision and remote sensing. \Cref{sec:dataset} describe the datasets employed in the study. \Cref{sec:methodology} presents the models, the metrics, and adopted preprocessing. \Cref{sec:results} presents the experimental results from both quantitative and qualitative perspectives. \Cref{sec:conclusion} draws the conclusion and presents future works. 

\section{Related Works}
\label{sec:rl}
In this section, we discuss the related works in canopy height estimation and monocular depth estimation.

\subsection{Canopy Height Estimation}
Various techniques have been employed to estimate canopy height, ranging from traditional field measurements~\cite{chm_onfield} to advanced remote sensing technologies~\cite{global_canopy_lidar}. One common tool is Light Detection and Ranging (LiDAR) due to its high spatial resolution and accuracy, allowing for detailed 3D representations of forest canopies~\cite{global_canopy_lidar}. The integration of LiDAR data with optical imagery and the application of machine learning algorithms have further enhanced canopy height estimation, thanks to the ability to deal with complex and heterogeneous environments~\cite{chm_lidar_satellite}.
The usage of LiDAR technologies is more expensive than RGB cameras, and generating canopy height maps (CHM) using deep learning models trained on remotely sensed satellite or aerial imagery has become popular as methods have been able to provide accurate and higher-resolution CHM estimates. Lang et al.~\cite{lang2023high} trained a model to produce a low-resolution 10m global canopy height map by fusing GEDI and Sentinel-2 satellite imagery. Becker et al.~\cite{becker2023country} similarly created a low-resolution map by using Bayesian deep learning to estimate canopy height in fused Sentinel-1 SAR and Sentinel-2 optical imagery. Pauls et al.~\cite{pauls2024estimating} improved upon these works by creating a novel shift resilient loss to adjust their ground truth for the intricacies of the GEDI global height product. Fogal et al.~\cite{fogel2024open} created a 1.5m resolution multitemporal dataset using aligned SPOT satellite imagery and ALS acquisitions in France. Tolan et al.~\cite{high_res_canopy_height_maps} took a self-supervised learning approach by pretraining a Vision Transformer (ViT)~\cite{dosovitskiy2021imageworth16x16words} using the DINOv2~\cite{dinov2} method on 18 million crops of globally sampled high-resolution WorldView satellite imagery. The authors then fine-tune this model to estimate canopy height using a dataset of U.S. based imagery and 1m resolution ALS-derived CHMs from the NEON catalog. While this approach works in practice, a significant amount of compute and efforts to collect these datasets are required.

\begin{figure}[htb]
    \centering
    \includegraphics[width=\linewidth]{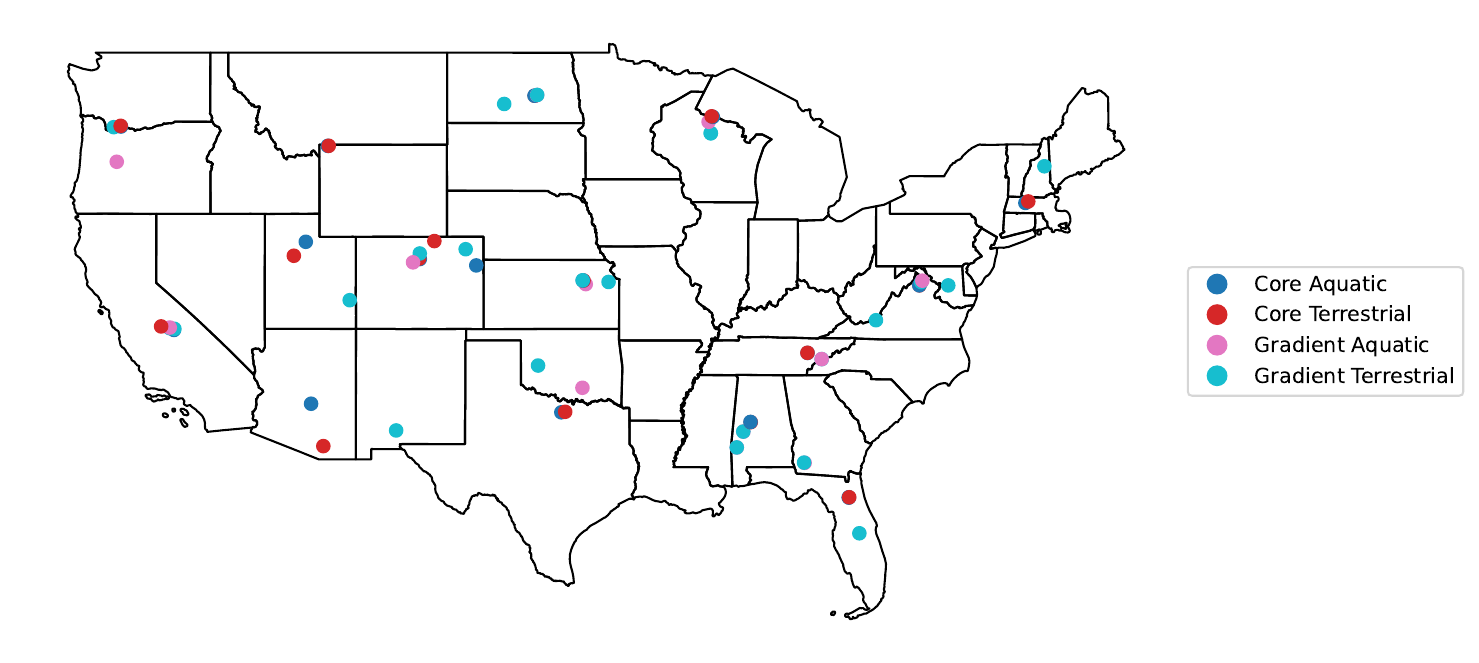}
    \caption{\textbf{National Ecological Observatory Network sites across the US.} Aquatic sites collect information about aquatic ecosystems, while terrestrial ones collect data about terrestrial ecosystems. Core sites provide long-term support, while Gradient sites are temporary sites to study the ecological response to specific changes.}
    \label{fig:neon_sites}
\end{figure}

\subsection{Foundation models in Computer Vision and Remote Sensing}
Foundation models are becoming one of the prevalent solutions in many fields when there is no sufficient training time or large data availability~\cite{foundation_history}, which is particularly relevant for transformers. In the computer vision field, ViT~\cite{dosovitskiy2021imageworth16x16words}, and subsequent CLIP\cite{clip} and DINOv2\cite{dinov2} have become well-established and robust encoders employed in many architectures and downstream tasks. Models like Segment Anything~\cite{sam} and OWL-ViT~\cite{owl_vit} make use of these advancements for general-purpose usage. 
In the remote sensing field, following these advancements, many solutions were proposed providing generative models~\cite{diffusionsat}, multimodal models~\cite{geochat}, or task-specific solutions~\cite{segmentanychange,ss_mae}. With this said, foundation models remain unexplored for the task of canopy height estimation.

\subsection{Monocular Depth Estimation}
Monocular depth estimation for natural imagery, being a more mainstream area of computer vision research, experiences faster improvements than remote sensing height estimation topics. The MiDaS/DPT family of models~\cite{lasinger2019towards,dpt,birkl2023midas} has shown that training for relative depth on a mixture of datasets, instead of training on small individual metric depth datasets such as NYU Depth V2~\cite{silberman2012indoor}, ETH3D~\cite{schops2017multi}, and DIODE~\cite{vasiljevic2019diode}, experiences better transfer learning performance. MegaDepth~\cite{li2018megadepth} and Depth Anything~\cite{depth_anything_v1, depth_anything_v2} have found that pretraining for relative depth estimation on higher quality and large-scale synthetic and pseudo-labeled images further improves zero-shot generalization performance. 

Furthermore, Marigold~\cite{ke2024repurposing} displayed the efficiency of requiring a single consumer NVIDIA RTX 4090 GPU for fine-tuning image generation diffusion models pretrained on large-scale datasets for the task of depth estimation.
These advancements in natural imagery provide an interesting and promising alternative to the full training of models for canopy height estimation, and thanks to the pre-training, they provide solid baselines for predicting relative distances.

\section{Datasets}
\label{sec:dataset}

In this section, we describe the data sources and the datasets employed for the analysis.

\subsection{Data Sources}
The primary data source for high-resolution CHM is derived from the National Ecological Observatory Network (NEON) catalog~\cite{neon_airborne}. The network is composed of many sites across the U.S., as shown in \Cref{fig:neon_sites}. Terrestrial sites collect data to monitor changes in climate, surface-atmosphere interactions, biogeochemical processes, organismal populations, and habitat structure over the land, while aquatic sites collect data to monitor changes in freshwater and biogeochemical processes, organismal populations, and habitat structure in water sources. Core sites provide long-term support, while Gradient sites
are temporary sites to study the ecological response to specific changes.
In our specific case, the data consists of high-resolution, multi-spectral, and hyperspectral aerial imagery captured from aircraft over various ecological sites. Multi-spectral imagery captures data in several specific wavelengths, typically including visible and near-infrared bands. Hyperspectral imagery captures data in hundreds of contiguous spectral bands. This imagery provides detailed spatial data on vegetation, land cover, and water bodies, aiding in vegetation health assessment, species distribution, and biomass monitoring.

\begin{figure}[t!]
    \centering
    \includegraphics[width=0.5\linewidth]{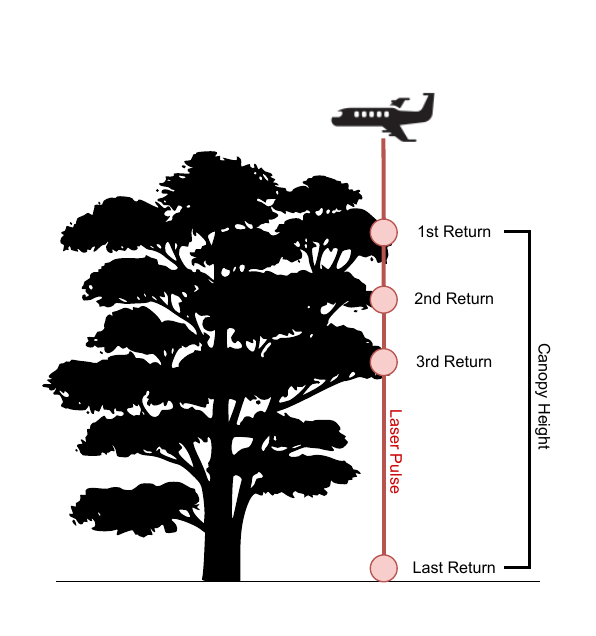}
    \caption{\textbf{ALS-derived tree canopy height acquisition process.} LiDAR sensors attached to a fixed-wing aircraft, emits a laser pulse that reflects many times until reaching the ground. The canopy height is computed using the time-delta between the first and the last return of the pulse, the pulse obtained when reaching the ground level.}
    \label{fig:chm_from_lidar}
\end{figure}

LiDAR sensors use laser light to measure distances to the Earth's surface and various objects from aircraft or drones. These sensors emit thousands of laser pulses per second toward the ground, capable of penetrating vegetation. These pulses hit various surfaces, such as the canopy, branches, and the ground, reflecting back to the sensor and providing punctual information. Points representing the highest surfaces in the forest canopy are used to create a Digital Surface Model (DSM), including the canopy's top. The Canopy Height Model (CHM) is derived by subtracting the Digital Terrain Model (DTM), which represents the bare earth surface, from the DSM. This provides the height of the canopy above the ground surface~\cite{lidar_vegetation_height} as shown in \Cref{fig:chm_from_lidar}.

\subsection{EarthView dataset}
The Satellogic EarthView Dataset~\cite{earthview} is a comprehensive collection of multispectral earth imagery for general-purpose tasks. The dataset is divided into four distinct subsets sourced from Satellogic, Sentinel-1, Sentinel-2 satellites and aerial imagery and ALS-derived CHMs from the NEON catalog. While the dataset provides satellite imagery provided by one commercial satellite (Satellogic) and two publicly available European Space Agency missions (Sentinel-1 and Sentinel-2), we employed only the aerial imagery, which is the only one with canopy height maps by LiDAR sensor, and it is the common imagery type between the two analyzed datasets. The NEON subset is composed of very high-resolution RGB images at 0.1m paired with 1m CHMs.

\subsection{High Resolution Canopy Height Maps dataset}
\label{sec:hrchm}
The High-Resolution Canopy Height Maps (HRCHM)~\cite{high_res_canopy_height_maps} is a collection of approximately 5,800 NEON CHMs with an area of 1 km $\times$ 1 km and a resolution of 1 meter per pixel. Samples were selected based on quality, minimal artifacts, and temporal proximity to a collection of Maxar satellite imagery. The CHMs were reprojected and resampled to match the Maxar imagery resolution and paired with a corresponding RGB satellite images to train a canopy height estimation model. The dataset was split into a train and test set, referred to as the \textit{NEON test set}, and preprocessed into 256 $\times$ 256 random crops.

\begin{figure}[t!]
    \centering
    \includegraphics[width=0.9\linewidth]{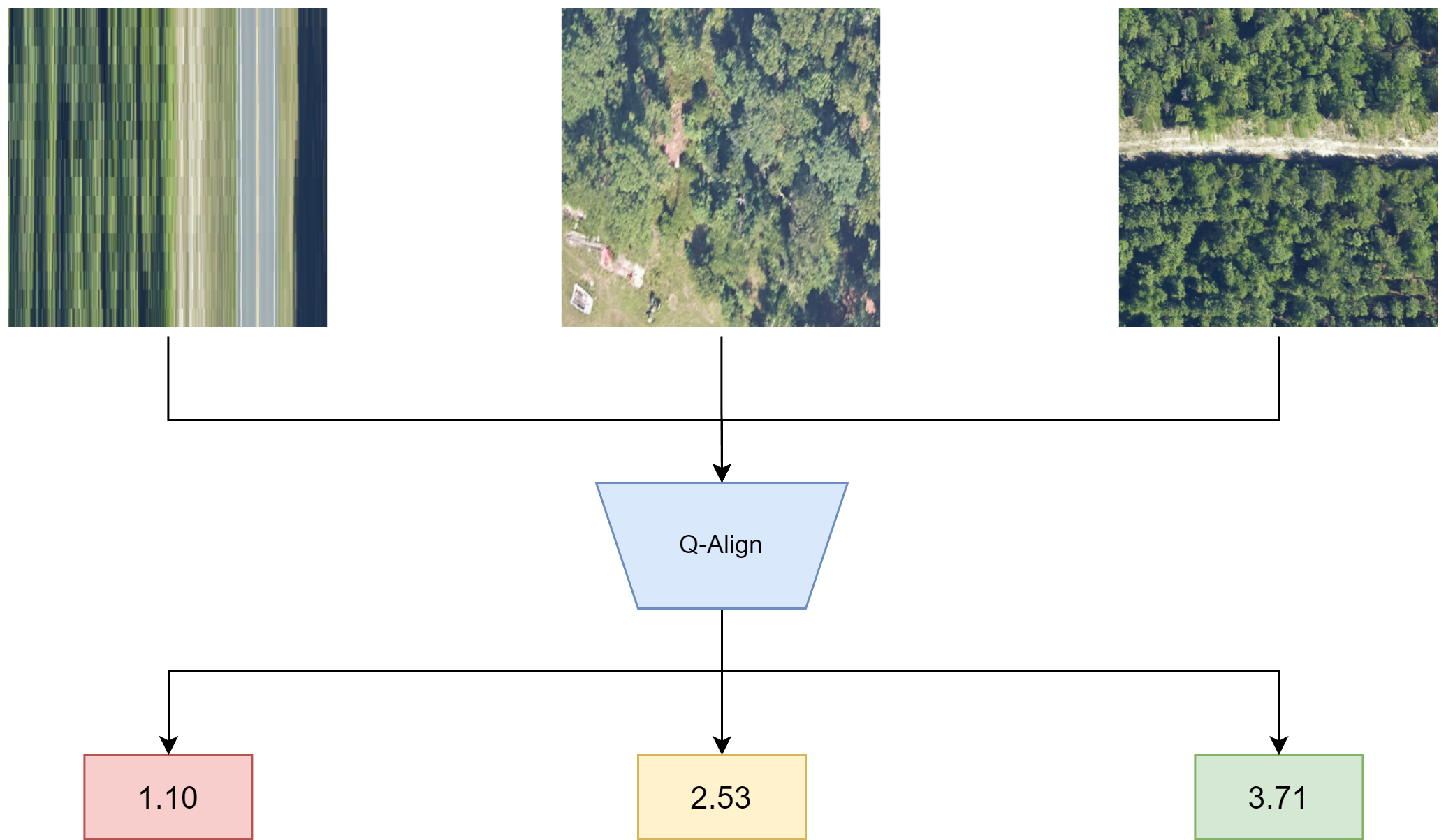}
    \caption{\textbf{Example quality scores by Q-Align~\cite{qalign} on NEON RGB images from the EarthView dataset.} On the left, a noisy sample scored \textit{1.10}; in the middle, a medium-quality sample scored \textit{2.53}; and on the right, a better-quality sample scored \textit{3.71}. This solution permits the detection and filtering of low-quality samples affected by warping and motion blurs.}
    \label{fig:qalign_example}
\end{figure}

\section{Methodology}
\label{sec:methodology}

This section outlines the model, preprocessing, and metrics used in the analysis.

\subsection{Problem Statement}
Let $I$ be an arbitrary RGB image of size $W \times H \times 3$, where $W$ and $H$ are the width and height of the image in pixels, respectively, and the channel depth of the image is 3, corresponding to the red, green, and blue color channels. The objective is to estimate the canopy height map $M$ from this image. $M$ can be represented by a matrix of size $W \times H$, where each element $M_{i,j}$ represents the estimated height of the canopy for the corresponding pixel $I_{i,j}$ in the image $I$.

\subsection{Preprocessing}
The NEON imagery in the EarthView dataset contains a significant number of samples with motion blur or warping due to image stitching errors. We find that these samples significantly degrade fine-tuning performance and are likely unuseful for evaluation. Therefore, we seek to filter these samples by employing the Q-Align~\cite{qalign} image quality assessment vision-language model to score the quality of each sample between a range of $[0-5]$ where higher values indicate better quality as shown in \Cref{fig:qalign_example}. We then filter low-quality samples using a score threshold of $t=2.5$. The training set contains 45781 samples, validation set 5788, and test set 5682. Every sample contains at least one pixel with the canopy height different from zero. According to KS-test the distribution of heights is similar between train-val ($p \approx 0.68$) and train-test ($p \approx 0.75$)

\begin{figure}[t!]
    \centering
    \includegraphics[width=0.9\linewidth]{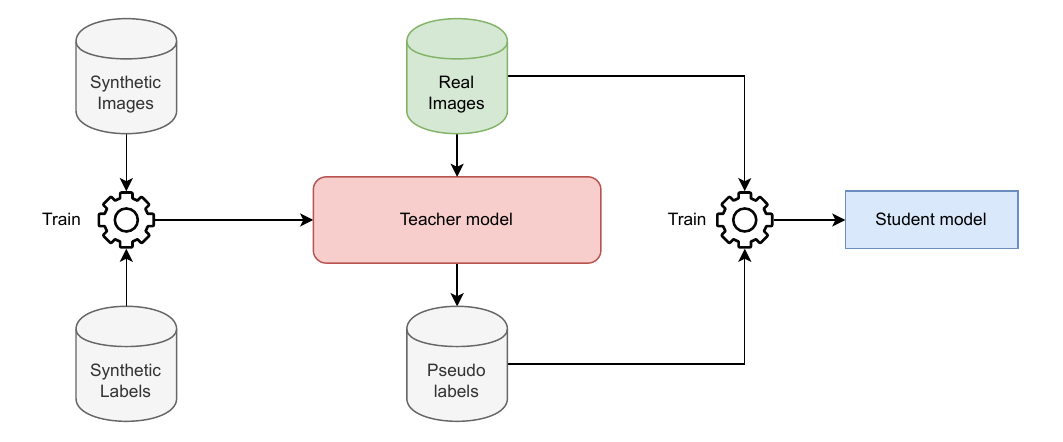}
    \caption{\textbf{Depth Anything v2 training procedure}. Synthetic images and relative labels are used to train a large teacher model. It is employed to annotate real images to create pseudo-labels. The real images and relative pseudo-labels are used to train a small student model.}
    \label{fig:depthanything_training}
\end{figure}

\subsection{Depth Anything for Canopy Height Estimation}
In our study, we used Depth Anything v2 in both zero-shot and with a finetuning on EarthView dataset, called Depth Any Canopy, for addressing the canopy height estimation.

Depth Anything~\cite{depth_anything_v1} is a state-of-the-art monocular depth estimation model. It features a DINOv2 pretrained ViT encoder~\cite{dinov2} and DPT decoder~\cite{dpt}. The model is trained using an online student-teacher distillation with a combination of real depth maps and pseudo-labeled depth maps of 62 million imagery. It excels in zero-shot generalization and improves performance when fine-tuned for metric depth on evaluation benchmark datasets. 

The successor, Depth Anything v2~\cite{depth_anything_v2}, takes a different approach and only pretrains on high-quality synthetic image and depth map pairs due to invalid pixels and low complexity of scenes in canonical metric depth datasets. It then uses the synthetic pretrained model to generate precise pseudo-labels of 62+ million real images in an offline manner. A student model is then trained on these pseudo-labels as shown in \Cref{fig:depthanything_training}. The model uses scale-and-shift invariant losses to ensure consistency and gradient matching loss for sharper depth predictions~\cite{dpt}. This model not only outperforms previous state-of-the-art models in accuracy, but also speed, and efficiency primarily due to the large-scale and high-quality pretraining dataset.

\begin{figure}[tb!]
    \centering
    \includegraphics[width=0.8\linewidth]{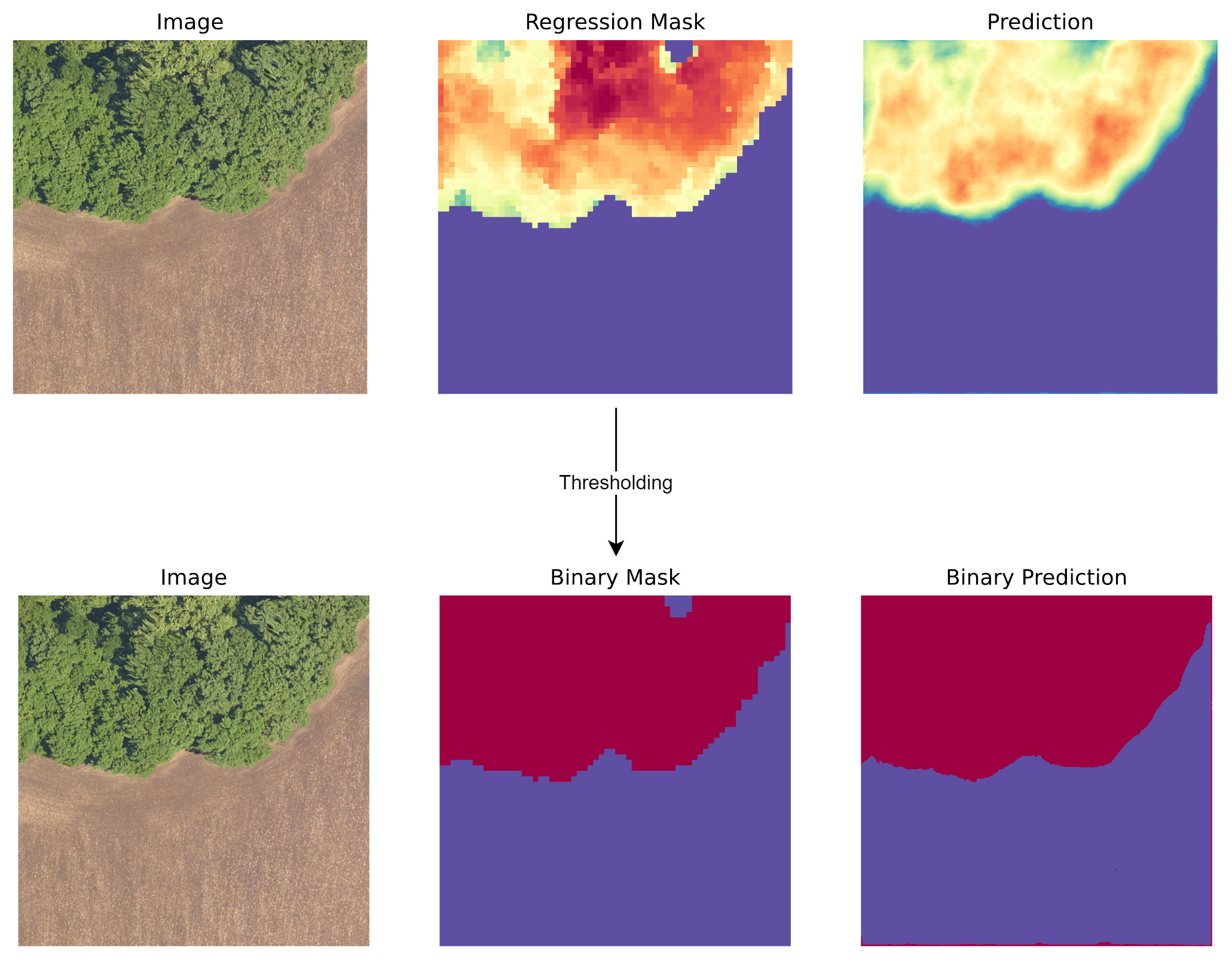}
    \caption{\textbf{Tree Canopy Extent evaluation process.} We threshold the ground truth and predicted canopy heights to obtain binary masks. We utilize this to evaluate a model's effectiveness in identifying tree extent.}
    \label{fig:iou_metric}
\end{figure}

\subsection{Baseline}
\label{sec:baseline}
As described in Section \ref{sec:rl}, Tolan et al.~\cite{high_res_canopy_height_maps} proposed to pretrain a model using the DINOv2 on a large-scale dataset of 18 million 256 $\times$ 256 RGB crops from global Maxar WorldView satellite imagery. This model was then fine-tuned on the HRCHM dataset described in Section \ref{sec:hrchm}.

\subsection{Metrics}

In our experimentation, we analyze the predicted performance on the canopy height estimation regression task using the Mean Absolute Error (MAE) (L1 distance). Furthermore, we evaluate the the ability of the model to segment the tree canopy from the background by using the same process by Tolan et al.~\cite{high_res_canopy_height_maps} to mask the predicted and ground truth CHM using a threshold of $t=1E-4$ as shown in \Cref{fig:iou_metric}. We then compute binary segmentation performance using the Intersection-Over-Union (IoU) metric to understand the quality of the predictions of areas considered as trees, tree canopy extent. Lastly we compute the Pearson Correlation (PC) of prediction and ground truth in the tree areas to understand whether relative heights are consistent despite the errors. To compare model size and efficiency, we report the floating-point operations per second (FLOPs) in billions (giga) and the number of parameters in millions.

\begin{table}[tb!]
    \centering
    \caption{\textbf{Results on EarthView and HRCHM datasets.} The best for each metric is bolded, while the second-best is underlined. \textit{FT} indicates if the model is finetuned on EarthView. \textit{DA} and \textit{DAC} refers to Depth Anything v2 and Depth Any Canopy, while \textit{DAC-S} and \textit{DAC-B} refers to the ViT-S and ViT-B variants, respectively.}
    \label{tab:results}
    \begin{tabular}{@{}l|c|cc|ccc|ccc@{}}
\toprule

\multirow{2}{*}{Model} &
\multirow{2}{*}{FT} &
\multirow{2}{*}{\# Params} &
\multirow{2}{*}{GFLOPs} &
\multicolumn{3}{c|}{EarthView\cite{earthview}} &
\multicolumn{3}{c}{HRCHM\cite{high_res_canopy_height_maps}} \\ 

\cline{5-10} 
\rule{0pt}{4mm}                      &                     &                            &                         & MAE $\downarrow$  & IoU $\uparrow$     & PC $\uparrow$     & MAE $\downarrow$  & IoU $\uparrow$     & PC $\uparrow$      \\ \midrule
SSL-H\cite{high_res_canopy_height_maps}                  & \xmark                   & 677M                       & 414                     & 0.2236   & 0.4164   & 0.1544   & \textbf{0.0306}  & 0.485  & \textbf{0.7441} \\
DA-S\cite{depth_anything_v2}                   & \xmark                   & \textbf{24.8M}                      & \textbf{115}                     & 0.4116   & 0.4164   & 0.2892   & 0.5960   & 0.6474 & 0.1791 \\
DA-B\cite{depth_anything_v2}                   & \xmark                   & 97.5M                      & 381                     & 0.4607   & 0.4164   & \textbf{0.361}    & 0.5972  & 0.6474 & 0.1692 \\ \midrule
DAC-S\cite{depth_anything_v2}                   & \tick                   & \textbf{24.8M}                      & \textbf{115}                     & \underline{0.1410}   & \underline{0.5323}   & 0.2740   & \underline{0.1025}  & \textbf{0.5672} & 0.6102 \\
DAC-B\cite{depth_anything_v2}                  & \tick                   & 97.5M                      & 381                     & \textbf{0.1304}   & \textbf{0.5926}   & \underline{0.3483}   & 0.1203  & \underline{0.5494} & \underline{0.6171} \\ \bottomrule
\end{tabular}
\end{table}

\section{Experimental Results}
\label{sec:results}

In this section, we present the experimental settings and an analysis of the results.

\subsection{Experimental Settings}
\label{sec:exp-settings}
The models were finetuned with the AdamW optimizer~\cite{loshchilov2017decoupled} for 3 epochs with a batch size of 8 on the EarthView training set using an NVIDIA RTX A6000 48GB GPU. We employed a learning rate scheduler with a warmup of 5\% of the training steps with linear decay. The maximum learning rate is set to $\alpha=5E-6$. We utilize the Mean Squared Error (L2 distance) between the ground truth and predicted canopy height as a loss function. The canopy height maps are min-max normalized to relative height.

We evaluate and fine-tune the Depth Anything v2 ViT-S and ViT-B checkpoints, referred to as DA-S and DA-B, respectively into Depth Any Canopy (DAC-S) and (DAC-B). We do not utilize the best performing ViT-G (Giant) weights because they have not been released. We baseline against the ViT-H (Huge) checkpoint from Tolan et al.~\cite{high_res_canopy_height_maps}, referred to as SSL-H, trained using the process described in Section \ref{sec:baseline}. We utilize this checkpoint because no other smaller ViT variants of this model were made available.

\textit{We note that the SSL-H baseline is comparatively larger than the Depth Anything v2 checkpoints we initialize from. Furthermore, SSL-H is already pretrained in-domain on 18 million remote sensing images and fine-tuned on a large set of NEON aerial RGB imagery for canopy height estimation. Therefore, the Depth Anything v2 weights should be at a significant disadvantage when comparing to this baseline.}

\subsection{Discussion}

\begin{figure}[tb!]
    \centering
    \begin{tabular}{cc}
        \rotatebox[origin=lt]{90}{\hspace{2mm} EarthView} & \includegraphics[width=0.9\linewidth]{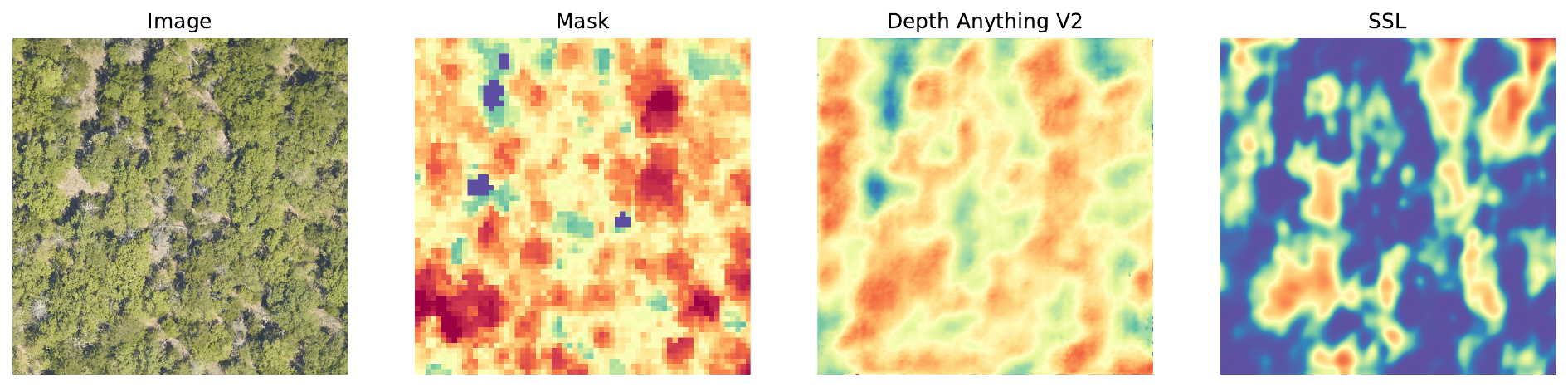} \\
        \rotatebox[origin=lt]{90}{\hspace{2mm} EarthView} & \includegraphics[width=0.9\linewidth]{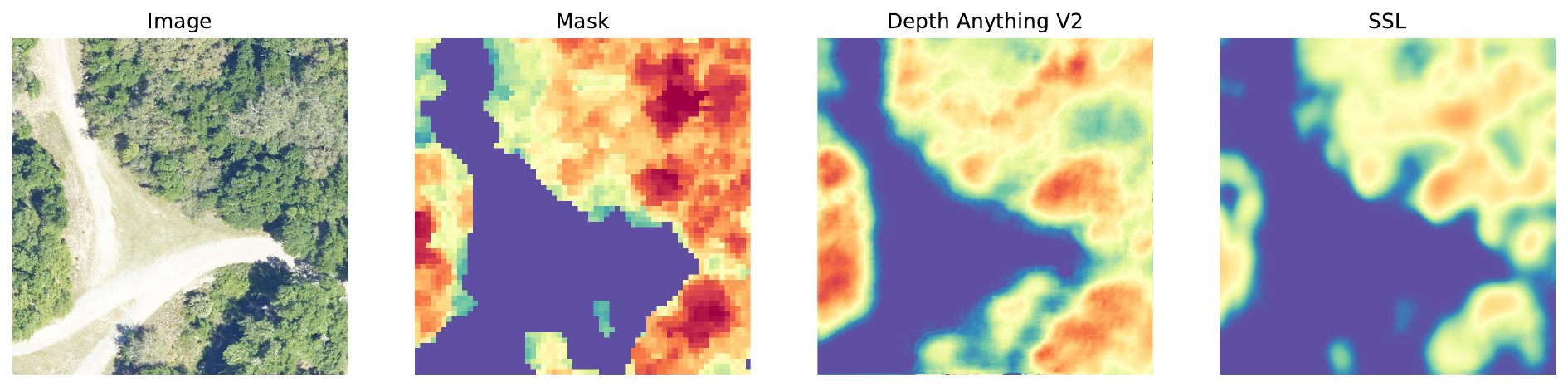} \\
        \rotatebox[origin=lt]{90}{\hspace{3mm} HRCHM} & \includegraphics[width=0.9\linewidth]{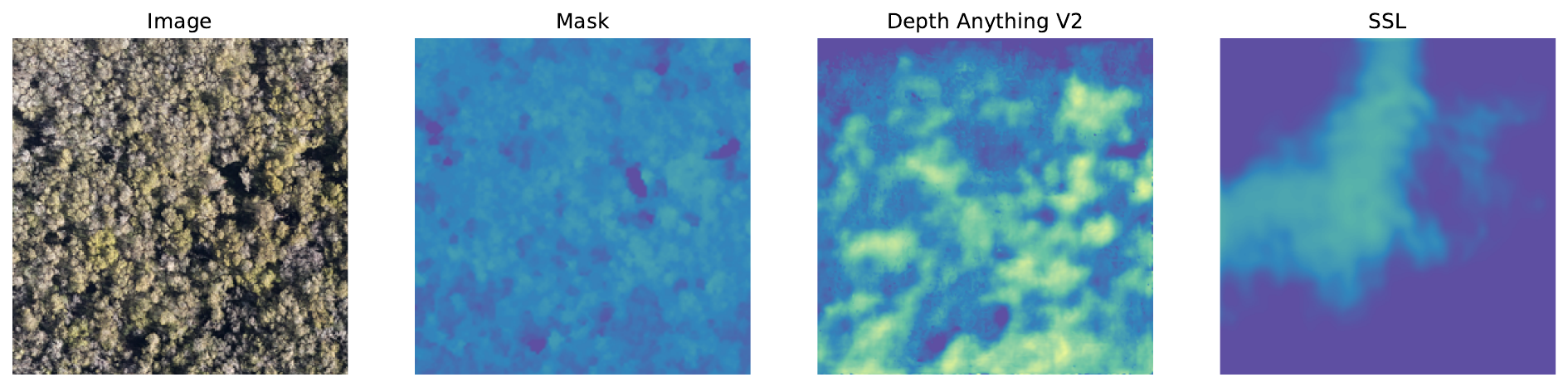} \\
        \rotatebox[origin=lt]{90}{\hspace{3mm} HRCHM} & \includegraphics[width=0.9\linewidth]{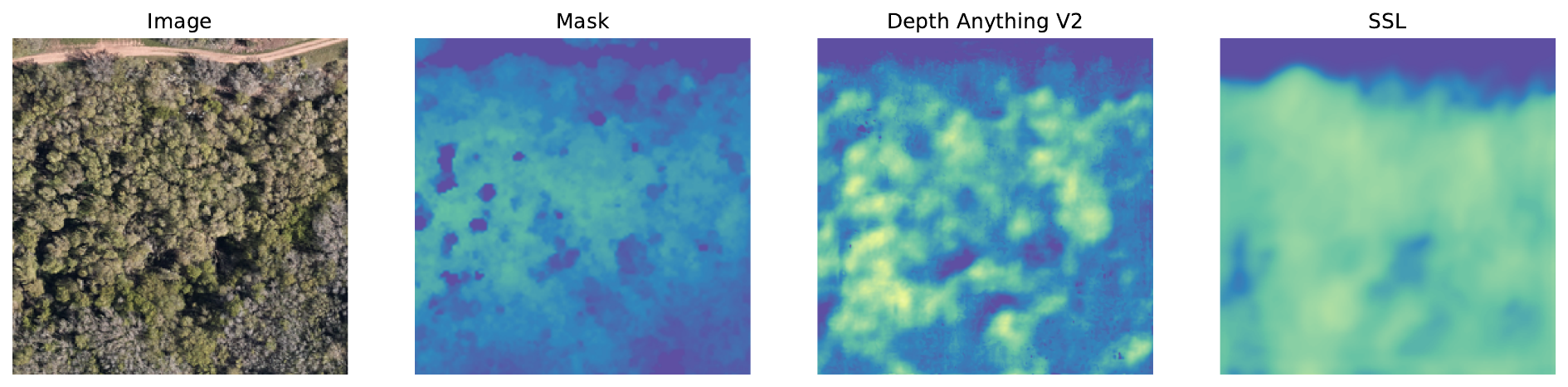} \\
    \end{tabular}
    \caption{\textbf{Example predictions of Depth Any Canopy (DAC-S) and SSL-Huge model from Tolan et al.~\cite{high_res_canopy_height_maps}} on NEON imagery from the EarthView and HRCHM datasets. Left to right: NEON RGB Image, Ground Truth Canopy Height Map, DAC-S predicted CHM and SSL-H predicted CHM.}
    \label{fig:example_predictions}
\end{figure}

\paragraph{\textbf{Overall Results}}
We tested the SSL-Huge (SSL-H)~\cite{high_res_canopy_height_maps} on both datasets (EarthView and HRCHM) to understand the performance and generalization capabilities. Depth-Anything (DA) was tested on both datasets in zero-shot and after finetuning on EarthView. This approach provides a comparison of both in-domain and out-of-domain data to understand the capability of generalization.

In \Cref{tab:results}, we reported the results obtained by each tested model. The fine-tuning of Depth Anything v2 on EarthView provides good comparative results and consistent performance under both MAE and IoU metrics, while the PC is generally higher on HRCHM for SSL-H.

SSL-H proves to be best in terms of MAE and PC on HRCHM, while the IoU is lower than the one achieved by DAC. The performance is worse on EarthView, particularly on MAE, which decreases by 10 times with respect to the performance on HRCHM. Additionally, we can reach good performance with smaller models.

\paragraph{\textbf{Zero-shot Comparison}}
When comparing Depth Anything v2 in zero-shot to Depth Any Canopy, we can conclude it is necessary to adapt the model to this new unseen task to be an effective solution. For example, shows a reduction in MAE from 0.4116 to 0.1410 on EarthView and from 0.5960 to 0.1025 on HRCHM after fine-tuning DA-S to DAC-S. However, the main advantages in terms of resource efficiency are maintained by providing good results in fewer GFLOPs. In many cases, DAC-S provides the best or second-best performance in around $1/4$ of the GFLOPs and $1/27$ of the parameters of SSL-H~\cite{high_res_canopy_height_maps}. DAC-S achieves results comparable to DAC-B, indicating it is a preferable solution for the task due to its low resource demands.

\paragraph{\textbf{Carbon Footprint Analysis}}
As mentioned in Section \ref{sec:exp-settings}, to achieve similar or comparable results to SSL-H by Tolan et al.~\cite{high_res_canopy_height_maps}, our DAC fine-tuning requires simply 3 epochs on the EarthView dataset using an NVIDIA RTX A6000 GPU. This requires 1.5 and 2.61 hours at an estimated cost of \$1.24 and \$2.09 using the \href{https://lambdalabs.com/service/gpu-cloud#pricing}{Lambda Labs} hourly pricing for the DAC-S and DAC-B variants, respectively. We calculate that our fine-tuning results in an estimated 0.14 and 0.24 kg of CO$_2$ emissions of using the \href{https://mlco2.github.io/impact#compute}{ML CO$_2$ Impact calculator}~\cite{lacoste2019quantifying}. We believe this to be preferable to the 8kg of CO$_2$ emissions reported by our SSL-H baseline for fine-tuning, not considering the 1.8T of CO$_2$ emissions for pretraining.

\paragraph{\textbf{Qualitative Results}}
\Cref{fig:example_predictions} provides a qualitative example of predictions by DAC-S and SSL-H on both datasets. On EarthView, SSL-H does not recognize many trees as expected, while DAC-S provides better performance. On HRCHM, we can note that SSL-H underestimates or overestimates some areas, although providing good IoU performance. The variations captured by DAC are generally superior to the ones captured by SSL-H.

\begin{figure}[tb!]
    \centering
    \subfloat[Example 1]{\includegraphics[width=0.9\linewidth]{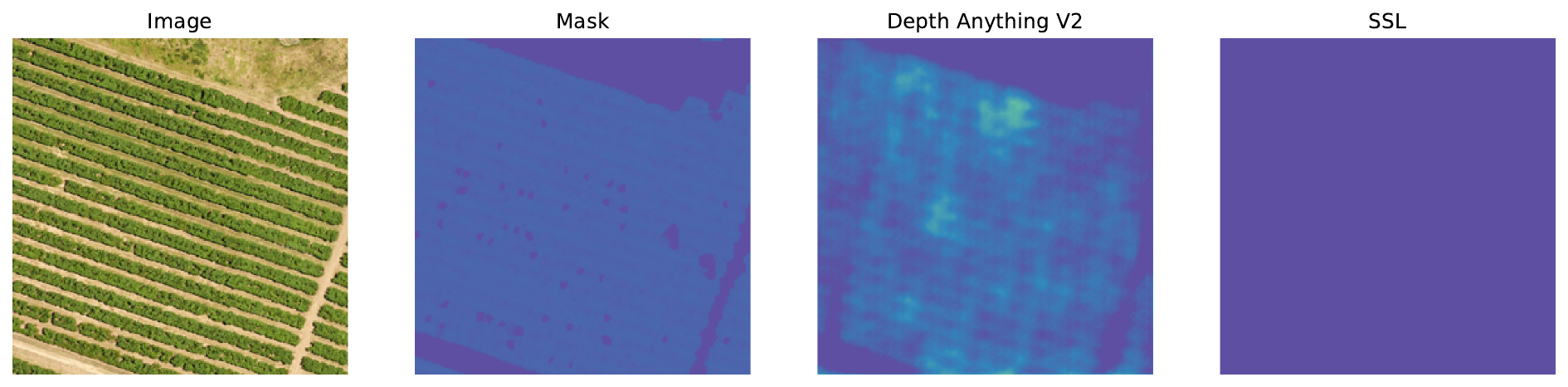}} \\
    \subfloat[Example 2]{\includegraphics[width=0.9\linewidth]{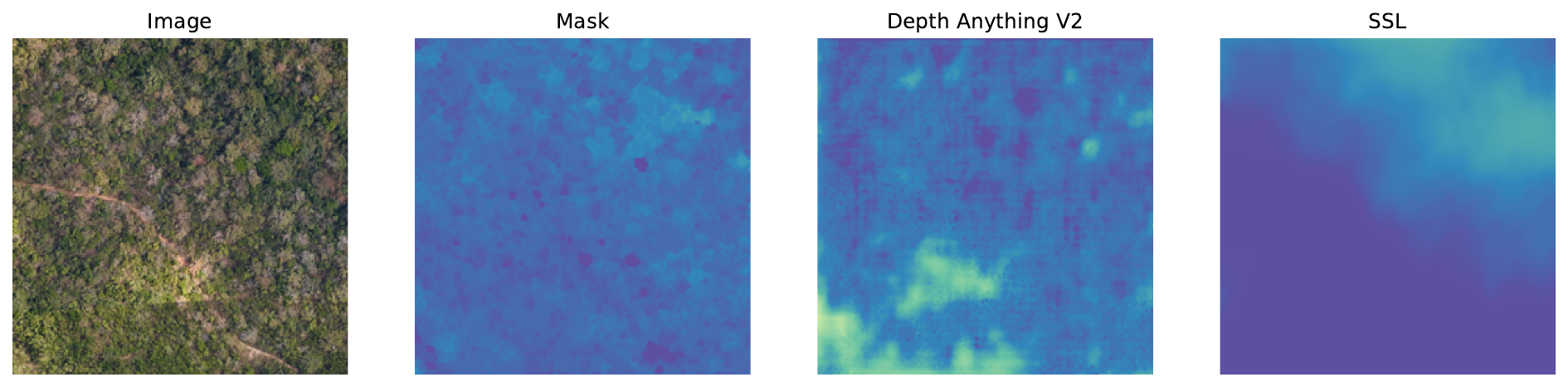}}
    \caption{\textbf{Examples of limitations of SSL-H addressed by Depth Any Canopy}. Through our analysis, we find that the SSL-H model by Tolan et al.~\cite{high_res_canopy_height_maps} tends to generate overly smooth predictions or predicts zero height for smaller vegetation. Because Depth Any Canopy is fine-tuned from the Depth Anything v2 weights it is able to recover vegetation heights for complex scenes and edge cases.}
    \label{fig:example_robustness}
\end{figure}

\Cref{fig:example_robustness} provides an overview of other cases. SSL-H avoids prediction on low trees, as in this qualitative example, while DAC-S tries to provide a height map of the area with a good level of detail. This shows the robustness achieved by Depth Any Canopy. The model can deal with complex scenarios, providing a high level of detail for complex scenes.

\section{Conclusion}
\label{sec:conclusion}
In this work, we have presented a novel approach to canopy height estimation by leveraging Depth Anything v2, a state-of-the-art monocular depth estimation foundation model. Our proposed model, Depth Any Canopy, demonstrates superior or comparable performance to current state-of-the-art methods with fewer computational resources. This is crucial for a scalable and cost-effective global canopy height estimation solution. 
Our findings show the potential of depth estimation foundation models pre-trained on large-scale natural imagery. They can be adapted for specific tasks in the remote sensing domain with minimal additional training. By fine-tuning Depth Anything v2, we have shown that it is possible to achieve high-quality canopy height maps from single-view images, overcoming the limitations of expensive pre-training on large-scale satellite imagery datasets to achieve comparable performance.
In future works, we plan to expand the evaluation to a wider variety of forest biomes and geographical regions to account for more diverse environments and to investigate the usage of hyperspectral and radiometric imagery, which could enhance the understanding of the area with complex features.

\bibliographystyle{splncs04}
\bibliography{refs}
\end{document}